\documentclass[10pt,twocolumn,letterpaper]{article}

\usepackage{cvpr}              

\usepackage{graphicx}
\usepackage{amsmath}
\usepackage{amssymb}
\usepackage{booktabs}
\usepackage{cite}
\usepackage{amsmath,amssymb,amsfonts}
\usepackage{algorithmic}
\usepackage{graphicx}
\usepackage{textcomp}
\usepackage{xcolor}
\usepackage{enumitem}
\usepackage{comment}
\usepackage{amsmath}
\usepackage{amssymb}
\usepackage{times}
\usepackage{epsfig}
\usepackage{multirow}
\usepackage[accsupp]{axessibility}  
\usepackage[pagebackref,breaklinks,colorlinks]{hyperref}

\usepackage[capitalize]{cleveref}
\crefname{section}{Sec.}{Secs.}
\Crefname{section}{Section}{Sections}
\Crefname{table}{Table}{Tables}
\crefname{table}{Tab.}{Tabs.}

\def\cvprPaperID{15} 
\def\confName{CVPR}
\def\confYear{2023}

\begin{document}

\title{BeCAPTCHA-Type: Biometric Keystroke Data Generation \\ for Improved Bot Detection}

\author{Daniel DeAlcala$^1$, Aythami Morales$^1$, Ruben Tolosana$^1$, Alejandro Acien$^1$, Julian Fierrez$^1$,\\
Santiago Hernandez$^1$, Miguel A. Ferrer$^2$, Moises Diaz$^2$\\
$^1$Biometrics and Data Pattern Analytics Lab, Universidad Autonoma de Madrid, Spain\\
$^2$University Las Palmas Gran Canaria, Spain\\
}
\maketitle

\begin{abstract}
   This work proposes a data driven learning model for the synthesis of keystroke biometric data. The proposed method is compared with two statistical approaches based on Universal and User-dependent models. These approaches are validated on the bot detection task, using the keystroke synthetic data to improve the training process of keystroke-based bot detection systems. Our experimental framework considers a dataset with $136$ million keystroke events from $168$ thousand subjects. We have analyzed the performance of the three synthesis approaches through qualitative and quantitative experiments. Different bot detectors are considered based on several supervised classifiers (Support Vector Machine, Random Forest, Gaussian Naive Bayes and a Long Short-Term Memory network) and a learning framework including human and synthetic samples. The experiments demonstrate the realism of the synthetic samples. The classification results suggest that in scenarios with large labeled data, these synthetic samples can be detected with high accuracy. However, in few-shot learning scenarios it represents an important challenge. Furthermore, these results show the great potential of the presented models.
\end{abstract}

\section{Introduction}
\label{sec:introduction}

The use of Artificial Intelligence (AI) in cyberattacks is an important concern for our society \cite{webcyber}. Along with the massive use of the Internet, the usage of bots to access digital services and platforms has grown, being the detection of these bots an open challenge with a high worldwide economical impact \cite{singh2019issues}. The rapid development of generative models during the last decade has allowed to synthesize realistic images and videos\cite{zhu2017unpaired,tolo_handbook_deepfakes}, audio \cite{wang2023neural}, or text data \cite{chatgpt}. These technologies can be integrated in new generations of bots with realistic human-like behavior. As an example, the Language Generation Models developed within the last two years have made almost impossible to distinguish between human and bot conversation. In this context, biometric technologies appear as a solution to distinguish between human and synthetic behaviors. 

 Biometric recognition is the ability to authenticate a person with the highest possible reliability based on their physical characteristics or behavioral attributes \cite{jain16}. This technology can be used to uniquely recognize one user among others (e.g., user identification), to recognize groups of subjects (e.g., soft-biometrics classification), or finally to differentiate real users from non-real users (e.g., bot detection). This work focuses on the topic of bot detection, more precisely in the generation and detection of synthetic keystroke patterns. Keystroke biometrics play an important role in bot detection due to its suitability in digital environments. Keyboards and touchscreens are among the most common human-machine interfaces nowadays, and their use in digital platforms and services is almost universal. 




In bot detection, a platform/system must detect bot attacks and differentiate them from legitimate user’s interactions. Traditionally, this detection has been carried out with conventional CAPTCHAS, which ask the user to perform some cognitive challenges. Most common conventional CAPTCHAS are: \textit{i)} Recognize characters in a distorted image; \textit{ii)} Identify a specific class in a set of images; and \textit{iii)} Analyze the interaction and web traffic.

Traditional CAPTCHA methods are becoming less and less effective due to advances in Computer Vision and the image classification approaches based on Deep Learning. As a result, other less intrusive and more effective CAPTCHAS are being developed nowadays based on the interaction information between the human/bot and the platform without actively requesting any information \cite{acien22mouse,acien21mobile}. The behavioral biometric characteristics, and specially the so called web-biometrics \cite{gamboa2007webbiometrics}, play an important role in this interaction modelling. These web-biometrics characteristics include keystroke dynamics, mouse dynamics, and mobile interaction, among others.

In this work we propose three synthetic keystroke data generation methods with application to bot detection. The main contributions of this work can be summarized as follow:

\begin{itemize}[noitemsep,topsep=0pt]
\setlength\itemsep{-0.0em}
    \item Three approaches for the synthesis of keystroke dynamics data based on Universal, User-dependent, and Generative Models. The first two proposed approaches are based on the statistical modelling of the biometric keystroke dynamics features of $100$,$000$ subjects. The third approach is based on a Generative Neural Network. In this work we demonstrate that data-driven learning approaches can be used to generate realistic keystroke dynamics, opening a new way to synthesize and detect keystroke samples.
    \item A bot detection method based on keystroke dynamics using algorithms trained with human and synthetically generated samples.
    \item A comprehensive performance analysis including: \textit{i)} amount of data available to train the bot detector; \textit{ii)} type of synthetic data used to model the human behavior; \textit{iii)} input text dependencies.
\end{itemize}

The rest of the work is organized as follows: Section 2 summarizes the related literature. Section 3 presents the proposed synthesis approaches. Section 4 describes the bot detection method. Section 5 presents the experimental results of the bot/human classification methods trained with the synthetic and human samples. Finally, in Section 6 we present the conclusions and limitations.

\vspace{3mm}
\section{Related Literature}

Keystroke dynamics has been widely focused on user recognition (i.e., differentiate one user from others). In 1980, a pioneer study of this biometric trait was made demonstrating that it is possible to differentiate subjects according to their typing patterns \cite{gaines1980}. In general, keystroke biometrics are commonly divided into two different approaches \cite{bergadano2002user}: free-text and fixed-text. Fixed-text approaches usually outperform free-text ones in terms of performance due to its lower intra-class variability. Nevertheless, the transparency and no restrictions of free-text approaches represent a clear advantage in most applications. 

During the last decades, the performance of keystroke biometric user recognition approaches has improved  to reach the actual state of the art. Some of these approaches include non-elastic sample alignment (e.g., Dynamic Time Warping \cite{morales16keystroke}), scaled Manhattan distances \cite{monaco2016robust}, and statistical models (e.g., Hidden Markov Models \cite{ali2016keystroke}). The performance of these approaches varies depending on the characteristics of the database and experimental protocol, but in general, Equal Error Rates (EER) over $5\%$ were consistently reported. During those years, the performance of free-text approaches was far from the performance achieved by fixed-text methods \cite{murphy2017shared,ayotte2020fast}. More recently, the release of new large-scale datasets and the use of Deep Neural Networks have boosted the performance of free-text keystroke biometrics with EERs under $5\%$ \cite{acien22transactions,morales2022setmargin,2023_FG_KeystrokeTransformers_Giuseppe}. 

The improvement of keystroke biometric technologies opens the doors to new applications apart from the traditional user recognition. One of these applications is bot detection. Bot detection presents some differences with respect to user authentication. While user authentication approaches are developed to model user-specific characteristics, bot detection approaches model the general population characteristics. The final aim is to extract the characteristics of human’s keystrokes dynamics and differentiate them from bots. 


Before getting into bot detection, we must talk about the synthesis of keystroke data generated by bots. One of the first studies was presented more than 10 years ago in \cite{stefan12} creating a synthetic database with $20$ subjects using first-order Markov chains. An improved keystroke biometric attack generator was presented in \cite{monaco2015spoofing}, using a Linguistic Buffer and Motor Control model. The use of synthetic keystroke samples to study the vulnerability of keystroke biometric systems was also studied in \cite{gonzalez2022towards,mhenni2019vulnerability,migdal2018analysis, migdal2019statistical}. Those studies have proposed methods using higher-order contexts and empirical distributions to generate impersonation attacks (i.e., samples generated to confuse the identity of an specific user). The conclusions from previous studies suggest that it is possible to generate realistic keystroke data.

\begin{figure*}[t!]
\centering
\begin{centering}
\includegraphics[width=1\linewidth]{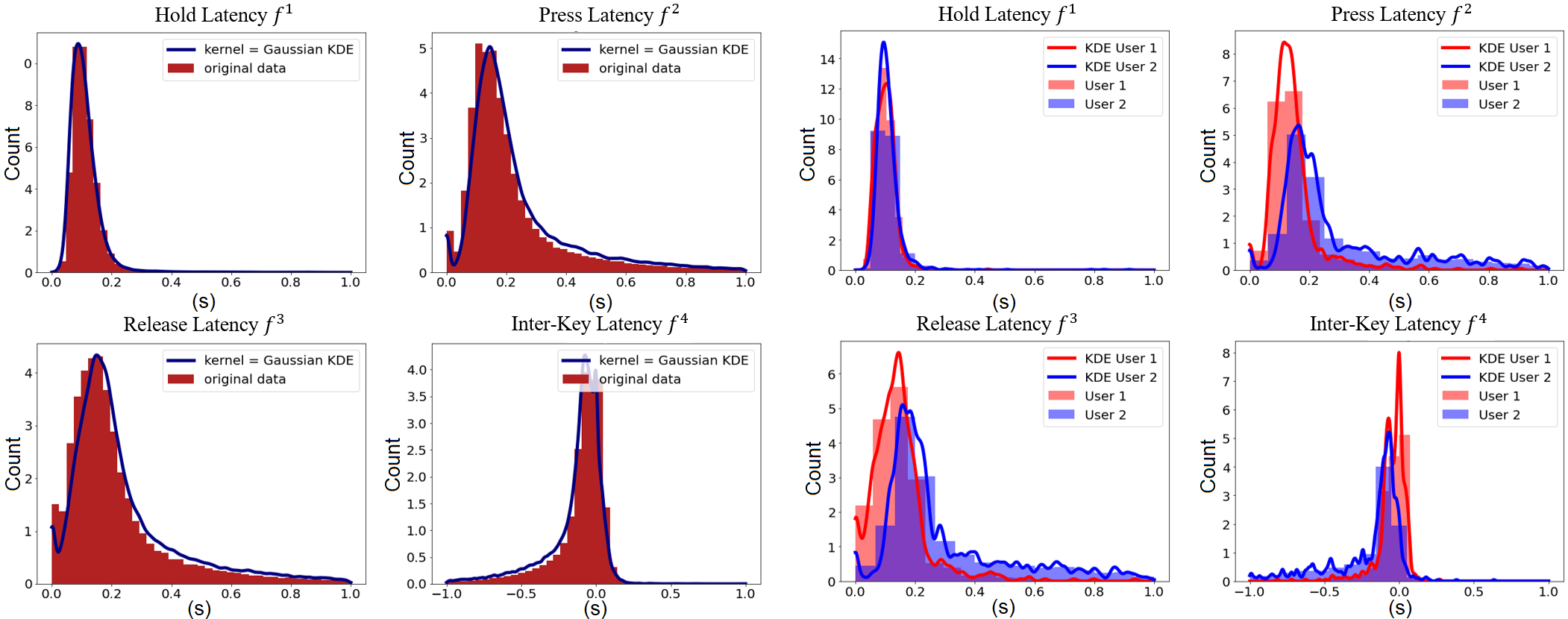}
\par\end{centering}
\caption[KDE]{Original human data (bars) and Kernel Density Estimator fitting model (continuous line, 4 images on the left) and the same for two independent subjects (4 images on the right).} 
\label{fig:KDEadjust}
\end{figure*}


Regarding keystroke bot detection itself, there is a pioneer work involving the use of function calls analysis \cite{al2008detecting}. The system proposed in \cite{al2008detecting} was based on communication protocol analysis (frequency of keyboard logs) rather than keystroke dynamics modelling. \textbf{Our work explores novel synthesis approaches based on keystroke dynamics for the development of new bot detection methods.} 

A lot of research is being done on keystrokes through statistical synthesis, creating synthetic samples that try to attack systems and that classifiers learn from \cite{gonzalez2022towards,mhenni2019vulnerability,migdal2018analysis, migdal2019statistical,alamri2022investigation,stefan12}. In this work we specially focus on papers \cite{alamri2022investigation} and \cite{stefan12} for the comparison with our proposed methods as they cover the topic of bot detection whereas the rest are more focused on the creation of synthetic samples and a very simplistic classification. In \cite{alamri2022investigation} the authors classified bots using Euclidean distance between human and bot features. In \cite{stefan12} they used a Support Vector Machine (SVM) classifier trained with real and synthetic samples. There is also a lot of research focused on the statistical synthesis of functions using Generative Neural Networks \cite{liu2021density,grover2018flow,uria2016neural}. In the present work we propose the use of traditional probability density estimators and a novel Generative Neural Network to improve the detection of keystroke synthetic data.

\subsection{Keystroke Dynamics Dataset}
\label{sec:database}
The Dhakal Dataset \cite{dhakal2018observations} is considered in this study to develop the models able to synthesize large-scale keystroke biometric data. There are $168$,$000$ subjects and 136 million of keystrokes in the database. Regarding the acquisition procedure, each subject had to learn a sentence and then write it as fast as possible (semi-fixed text scenario) using their own keyboard. Each subject has 15 sentences with a minimum of 3 words and a maximum of 70 characters.

Following the traditional keystroke dynamics modelling, the dataset is processed to extract 4 time features derived from the two main typing events (key press and key release) and the ASCII code for each key pressed \cite{acien22transactions}:
\begin{enumerate}
\setlength
    \item Hold Latency ($f^{1}_{j}$): Time between the key $j$ is pressed and released.
    \item Inter-Press Latency ($f^{2}_{j}$): Time between two consecutive keys are pressed, $j$ and $j+1$.
    \item Inter-Release Latency ($f^{3}_{j}$):  Time between two consecutive keys are released, $j$ and $j+1$.
    \item Inter-Key Latency ($f^{4}_{j}$):  Time between a key $j$ is released and the next key $j+1$ is pressed.
    \item Key Code ($f^{5}_{j}$): ASCII code normalized between 0 and 1 for each key $j$.
\end{enumerate}

Figure \ref{fig:KDEadjust} (left) presents the distribution of these features for the complete dataset. As can be seen, the Hold Latency feature follows a normal. The rest of the features presents tails related to the characteristics of the typist and the key-pressed (some combinations of keys used to present larger timing than others). Note that the Inter-Key Time feature presents negative times, i.e., the next key is pressed before the currently one is released. This effect is called rollover-typing and it is common in keystroke recognition systems.

\section{Keystroke Synthesis: General Outlook}
\label{sec:guidelines}

This section presents our three keystroke dynamic data synthesis methods. There are two different approaches: the first one is based on the statistical modelling of the feature distribution of the keystroke time series, while the second is based on a data-driven learning approach with a Generative Neural Network. As described in the previous section, the keystroke dynamic features model the biometric patterns during a typing task as differences of times (i.e., time gaps between key press and key release events). We propose to model the probability distribution of the keystroke biometrics features $\textbf{f}^{i}=[f^{i}_{0},...,f^{i}_{N}]$ to generate realistic human-like time sequences where $N$ is the total number of samples used. To train the following methods, the $N$ samples used are from 100,000 subjects out of the 168,000 in the database.

\subsection{Statistical Generative Models}
For the statistical approach we use the Kernel Density Estimator algorithm (\textit{KDE}) \cite{kim2012robust}. \textit{KDE} is a nonparametric algorithm that estimates univariate or multivariate densities. \textit{KDE} allows to compute the density of keystroke biometrics features as a set of functions $\mathrm{F}=[\mathrm{F}^1,\mathrm{F}^2,\mathrm{F}^3,\mathrm{F}^4]$, here the \textit{KDE} approximates each one of the time features $\textbf{f}^{i}$ at a point $x$ as:

\begin{equation}
    \label{eqn:kde}
    \mathrm{F}^i(x)=\frac{1}{N}\sum^{N}_{j=1}K(x-f^{i}_j;\sigma)
\end{equation}

\noindent where $K$ is the kernel function (Gaussian in our experiments) and $\sigma$ is the bandwidth ($\sigma=1.0$ in our experiments). We use this method to model the probability distributions $\mathrm{F}$ of the keystroke biometric features in the Dhakal Dataset (see Figure \ref{fig:KDEadjust} continuous lines). The synthesis of keystroke dynamic samples is divided into: 1) generation of a sequence of $K$ keys representing the typed text: $\textbf{k}=[k_0,...,k_K]$; 2) generation of the corresponding keystroke biometric features $\textbf{f}^{i}$ as random samples from the the learned models $\mathrm{F}$ (random sampling serves to introduce human-like variability between samples); and 3) the calculation of a sequence of timestamps: $\textbf{t}^\prime=[t^\prime_0,...,t^\prime_{2 \times K}]$ associated to the key press and key release events. The timestamp vector $\textbf{t}^\prime$ can be easily obtained from the time features $\textbf{f}^{1}$ and $\textbf{f}^{4}$. The following equations show the calculation of timestamps for the first two keys: 
\begin{align}
\begin{split}
\label{eqn:time_stamps_0}
    t^\prime_0=0, t^\prime_1=t^\prime_0+f^{1}_{1} \rightarrow \textrm{(key $1$)} 
\end{split} \\ 
\begin{split}
\label{eqn:time_stamps_1}
    t^\prime_2=t^\prime_1+f^{4}_{1} , t^\prime_3=t^\prime_2+f^{1}_{2} \rightarrow \textrm{(key $2$)}
\end{split}
\end{align}

\noindent The values of $\textbf{f}^{1}$ and $\textbf{f}^{4}$ are generated using the \textit{KDE} functions $\mathrm{F}^1$ and $\mathrm{F}^4$. We propose two synthesis approaches depending on the information used to model the feature distributions of $\textbf{f}^{i}$: Universal Model or User-dependent Model.

\subsubsection{Statistical Approach 1: Universal Model}
{The Universal Model is based on the estimation of a unique set of \textit{KDE} functions $\mathrm{F}$ representing the behavior of all subjects in the Dhakal dataset. As a result, only 4 \textit{KDEs} are necessary to model the human typing behavior distributions. Figure \ref{fig:KDEadjust} shows the set of trained functions (continuous lines in the four images on the left). In general, this approach could approximate in a good way the human features as a group. However, it could also generate unnatural samples due to the combination of every subjects times. It is important to highlight that this universal synthesis approach is not able to model: the intra-user dependencies (i.e., each user has certain biometric features and a correlation between them) or the key-dependent features (i.e., each key has a typing pattern depending on itself and also to a certain extent on the previous and following keys). First, the keystroke timestamps  $\textbf{t}$ from real human samples is parameterized according to the four time features $\textbf{f}^{i}$. Second, the probability function $\mathrm{F}^i$ of each time feature is independently modeled according to a \textit{KDE} function (see Eq. \ref{eqn:kde}). The four trained \textit{KDE} are then used to generate new keystroke dynamic features $\textbf{f}^\prime$ from which the synthetic keystroke timestamps $\textbf{t}^\prime$ are obtained (see Eq. \ref{eqn:time_stamps_0} and \ref{eqn:time_stamps_1}).}

\subsubsection{Statistical Approach 2: User-dependent Model}

The fundamental principle of keystroke biometric recognition systems is that typing patterns vary from one user to another. The User-dependent generation method tries to incorporate this intra-user characteristics into the synthesis process. However, the data available for each user is limited and therefore, the User-dependent Models could be less accurate than the previous Universal Model. Thus, the user-dependent synthesis approach is aimed to capture the relations between the dynamics features $\textbf{f}^{u,i}$ along the sentence typed by the user $u$. Even so, with this model we are still not able to model the key-dependent features. First, keystroke samples from the Dhakal dataset data are divided by subjects. Second, the keystroke timestamps data $\textbf{t}^{u}$ from user $u$ is parameterized to obtain the four time feature sequences $\textbf{f}^{u,i}$. Then, the probability distribution of each time feature from each training user ($\mathrm{F}^{u,i}$) is independently modeled according to a \textit{KDE} function (i.e., four $\mathrm{F}^{u,i}$ per user). This process is repeated for $M$ different human subjects in the database. Finally, the $M$ models are used to generate synthetic feature vectors $\textbf{f}^{u\prime}$ and its corresponding synthetic keystroke timestamps $\textbf{t}^{u\prime}$ following the probability distributions of independent human subjects.

\subsection{Generative Neural Network Model}

We propose a novel Generative Neural Network (GNN) for the synthesis of keystroke time series. The synthesis of time series with random distributions using Generative Neural Networks is an open challenge in the literature \cite{liu2021density,grover2018flow,uria2016neural}. If we want the GNN to learn a distribution we can not use a normal loss function because they are not optimized to learn distributions. For example, if we use the regression function Mean Square Error (MSE) for each different key-code, the network will learn to give always the same value (deterministic output), the one that appears more in the distribution to minimize losses (e.g., in a Gaussian distribution the output will be the mean). 

\begin{figure}[t!]
\begin{centering}
\includegraphics[width=\columnwidth]{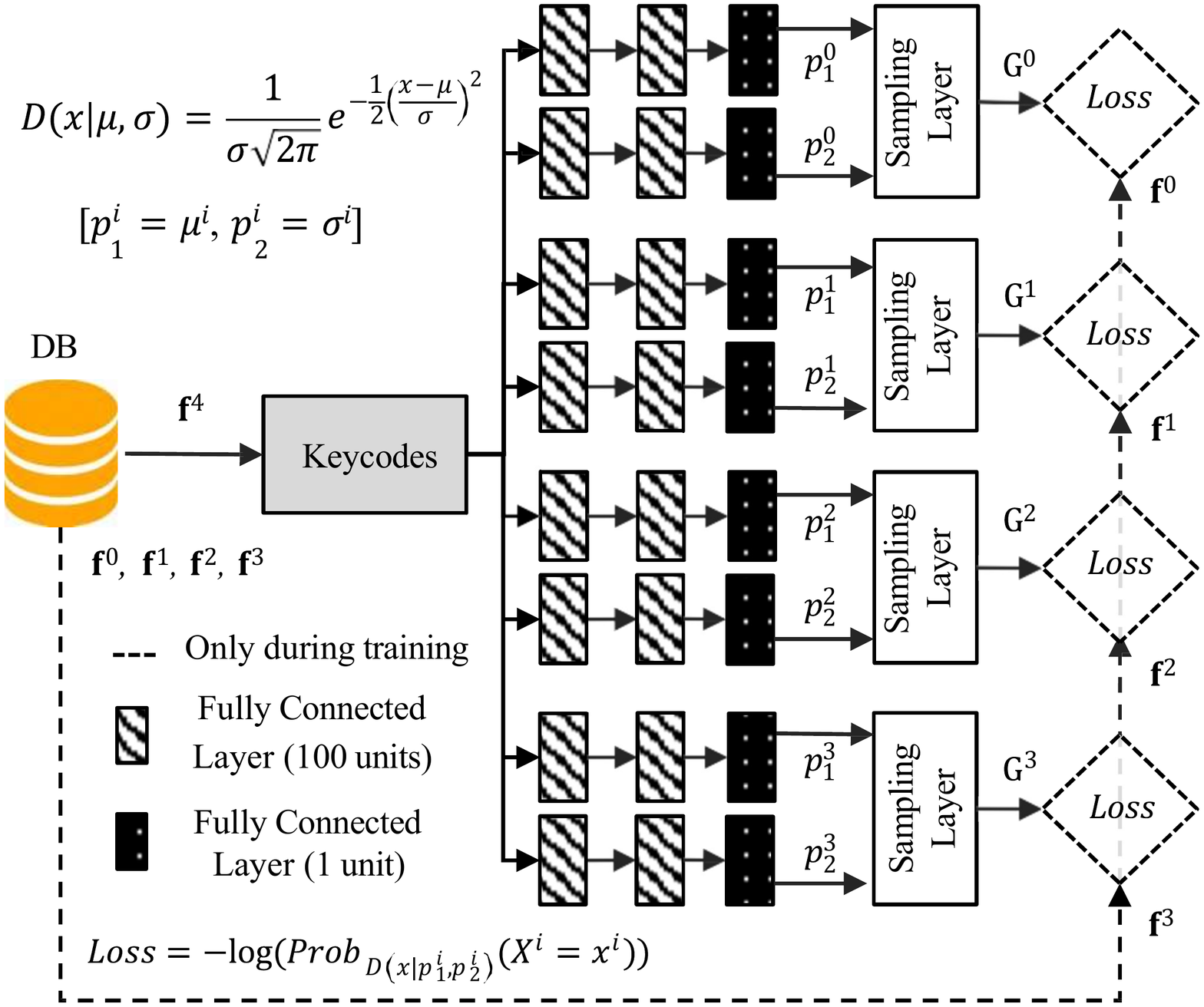}
\par\end{centering}
\caption[Created GNN]{Proposed Generative Neural Network learning framework. Example based on a Gaussian distribution defined by $\mu$ and $\sigma$ parameters.} 
\label{fig:GNN}
\end{figure}

The aim of our GNN model is to learn the required parameters to synthesize the realistic keystroke biometrics features $\textbf{f}^{i}_{j}$ with realistic intra-class and inter-class variability intrinsic to all biometric data. The input of the model is a key-code and the GNN generates different human-like times for this specific key-code. We train a specific GNN model for each time feature ($\textbf{f}^{i}$) thus obtaining a set of 4 functions $\mathrm{G}=[\mathrm{G}^0,\mathrm{G}^1,\mathrm{G}^2,\mathrm{G}^3]$. The GNN could be described as a function as shown below:

\begin{equation}
    \label{eqn:GNN}
    \mathrm{G}^i(k)=GNN(k;W^i)
\end{equation}

\noindent where $k$ is the key code and GNN is a neural network with its weights ($W^i$). During the training process the key-codes are introduced as input, and the network learns the time distributions for each $\textbf{f}^{i}$ from the real data. During the inference process, the code is introduced as input and the networks generate the $4$ time features. The proposed GNN computes the required parameters for the distribution of each key-code and then randomly samples this distribution. Figure \ref{fig:GNN} shows an example of our proposed GNN learning framework based on a Gaussian distribution with parameters $\mu$ and $\sigma$. The proposed architecture is based on: two fully-connected layers with $100$ units each (tanh activation function), one fully-connected layer with $1$ unit (linear activation function) and one sampling layer (this layer creates the probability density function with the output of the previous layer and samples it). 

The training process of the GNN is designed to learn the parameters of the statistical distributions of the features $\textbf{f}^{i}$. In our experiments, each time feature ($\textbf{f}^{i}$) is modelled by a parametric function defined by $q$ parameters ($\textbf{P}^i=[p^{i}_1,...,p^{i}_q]$). For example, a Gaussian distribution can be modelled by two parameters: mean and variance ($\textbf{P}^i=[\mu^i,\sigma^i]$).  The loss function used computes the probability of each sample (${X^i}$) to belong to the distribution ($D$) generated with the parameters $\textbf{P}^i=[p^{i}_1,...,p^{i}_q]$ learned up to that training moment:

\begin{equation}
    \label{eqn:loss}
    Loss = -\log{(Prob_{D(\textbf{x$\mid$ P}^i)}(X^i=x^i))}
\end{equation}

The loss value is the probability of the real time (the time used for training, $\textbf{f}^{i}$) to belong the distribution with the parameters learned up to that training point. This loss function acts as the Likelihood Function of the distribution we want to learn. 

For the proposed Generative Neural Network we only consider a Universal Method since for a User-dependent one we would need a large number of samples from each subject (the Dhakal database is large in number of subjects but not in samples per subject). Finally, to generate the time series associated to an specific sequence of key-codes, we employed the learned functions $ \mathrm{G}$ and the equations presented in Eq. \ref{eqn:time_stamps_0} and \ref{eqn:time_stamps_1}.

\section{Keystroke Bot/Human Classification Enhanced with Synthetic Data}
\label{sec: Bot_Human_Classification}
Most bots are not developed to generate realistic keystroke time series. They are usually developed to interact with a web service/platforms and this interaction usually includes introducing text as input (e.g., searching information) \cite{singh2019issues}. The code of a traditional bot is exclusively focused on the generation of a sequence of keys $\textbf{k}$ necessary to produce a desired result. This work explores a more challenging scenario where the bot is developed to spoof a keystroke bot detection system,  generating human-like keystroke time sequences $\textbf{t}^\prime$.

\begin{figure*}[t!]
\begin{centering}
\includegraphics[width=0.80\linewidth]{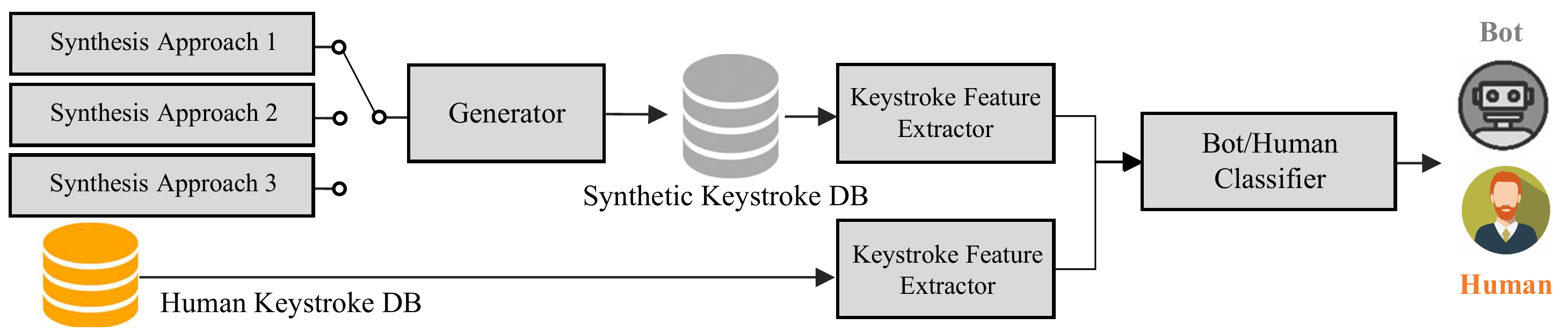}
\par\end{centering}
\caption[bot_detection]{Application of the three proposed keystroke data synthesis approaches to bot detection.} 
\label{fig:bot_detection}
\end{figure*}

We propose the use of synthetic keystroke samples to train bot detection systems (see Figure \ref{fig:bot_detection}). We use the Dhakal Dataset \cite{dhakal2018observations} to model the real human keystroke patterns in our experiments (see Section \ref{sec:database} for details about the dataset). First, the synthetic samples are generated using the text from the real ones, i.e., the human and bot key sequences are exactly the sames so that the classifier cannot differentiate them by the key codes. Note that, the Dhakal Dataset was captured according to a semi-fixed text protocol. This protocol implies that the text varies for each human sample in most of the cases. Second, the human and synthetic keystroke sequences are truncated to $L$ ($L$ is equal to $30$ in our experiments) or if smaller they are discarded. The reason for truncating these sequences is to be able to use the different classifiers that are presented below, under the same conditions. Third, each keystroke sequence is parameterized according to the features $f^{i}$ presented in Section \ref{sec:database}. Finally, we evaluate four different classification algorithms: Support Vector Machine (SVM), Recurrent Neural Network based on Long Short-Term Memory (LSTM), Random Forest (RF) and Gaussian Naive Bayes (GNB). These algorithms are trained using both human and bot feature vectors. We describe in Section \ref{sec:protocol} the experimental protocol details. These four classifiers have been chosen, since they are some of the most relevant and have been developed under very different applications, which will allow us an in-depth analysis.


There could be a question whether the use of synthetic samples to train the classifiers is really useful and improves their performance or not. To shed some light on this, we introduced a One-Class Classifier (One-Class SVM) trained exclusively with human samples. With this result it is possible to compare whether, at least in this type of classifier (although it can certainly be extrapolated to the rest), it is being useful to include these samples. This One Class training is not carried out with the rest of the classifiers because there is no implementation or theoretically it does not make sense.

\section{Experiments and Results}

\subsection{Experimental Protocol}
\label{sec:protocol} 
We divide the Dhakal database in two sets with $100$,$000$ subjects for training and the remaining $68$,$000$ subjects for test. Including or not the key-codes in the classifier can have great relevance in the detection of bots since one of our synthesis approaches takes into account the key for the creation of times whereas the others not. As a result, we consider experiments with and without the key-codes to better understand its impact in the performance. In addition, Privacy-preserving have a great importance in our society \cite{2022_ACM_Privacy_Paula_Giuseppe} and therefore the use of key-codes may not be available to protect access to the information typed by the subjects. It is therefore interesting to see the performance of the classifiers when this information is available or not.

The scarcity of labeled bot samples is a common challenge in bot detection \cite{singh2019issues}. For this reason, the experiments are divided into different scenarios depending on the data available to train the bot detector (from $20$ subjects to $500$). Each subject contains 15 real and 15 synthetic samples. Therefore, training with $20$ subjects result in $300$ synthetic samples and $300$ human samples. All the models are evaluated using the same $500$ bot and $500$ human samples ($15$,$000$ samples in total).

Two main objectives are considered in the analysis. First, the quantitative evaluation of the synthesis methods (\textbf{O1}). Second, the evaluation of the keystroke bot detection (\textbf{O2}). Previous objectives (\textbf{O1}, \textbf{O2}) are analyzed depending on the multiple variables mentioned: \textit{i)} the number of bot samples available to train the classifier; \textit{ii)} the classification algorithm (LSTM, SVM, RF and GNB) and \textit{iii)} the availability of key-codes to train the classifiers.



The experimental protocol comprises two scenarios assuming the availability or not of synthetic data: \textit{i)} Closed Set (Table \ref{Table:ResultsClass}); and \textit{ii)} Open Set (Table \ref{Table:ResultsClassOS}). In the Closed Set scenario we trained the bot detector with samples generated with the same synthesizer used in test (different samples but same generation method). In the Open Set scenario we train the detector using samples generated with a synthesizer different to the one used for the test samples. This scenario allows to evaluate the generalization capacity of the bot detectors. Everything is evaluated according to the classification accuracy of the model (i.e., human/real vs synthetic/bot). 

It is important to note that the aim of the experiments is not to confront the different generation approaches but to analyze how they can be used to detect bots generated with different synthesis approaches. The final aim is to generate synthetic databases that can be used to train these bot detectors.

\begin{table*}[t!]
\centering
\begin{tabular}{ll|llllllllll|}
\cline{3-12}
 &  & \multicolumn{10}{c|}{Clasification Model} \\ \cline{3-12} 
 &  & \multicolumn{2}{c|}{OC SVM} & \multicolumn{2}{c|}{SVM} & \multicolumn{2}{c|}{GNB} & \multicolumn{2}{c|}{RF} & \multicolumn{2}{c|}{LSTM} \\ \hline
\multicolumn{1}{|l|}{Gen Model} & \# Train Subjects & \multicolumn{1}{l|}{K=0} & \multicolumn{1}{l|}{K=1} & \multicolumn{1}{l|}{K=0} & \multicolumn{1}{l|}{K=1} & \multicolumn{1}{l|}{K=0} & \multicolumn{1}{l|}{K=1} & \multicolumn{1}{l|}{K=0} & \multicolumn{1}{l|}{K=1} & \multicolumn{1}{l|}{K=0} & K=1 \\ \hline
\multicolumn{1}{|l|}{\multirow{3}{*}{User-dep}} & 20 & \multicolumn{1}{l|}{0.43} & \multicolumn{1}{l|}{0.44} & \multicolumn{1}{l|}{0.79} & \multicolumn{1}{l|}{0.77} & \multicolumn{1}{l|}{0.65} & \multicolumn{1}{l|}{0.65} & \multicolumn{1}{l|}{0.87} & \multicolumn{1}{l|}{0.88} & \multicolumn{1}{l|}{0.50} & 0.50 \\ \cline{2-12} 
\multicolumn{1}{|l|}{} & 100 & \multicolumn{1}{l|}{0.54} & \multicolumn{1}{l|}{0.54} & \multicolumn{1}{l|}{0.83} & \multicolumn{1}{l|}{0.79} & \multicolumn{1}{l|}{0.63} & \multicolumn{1}{l|}{0.63} & \multicolumn{1}{l|}{0.92} & \multicolumn{1}{l|}{0.92} & \multicolumn{1}{l|}{0.51} & 0.51 \\ \cline{2-12} 
\multicolumn{1}{|l|}{} & 500 & \multicolumn{1}{l|}{0.53} & \multicolumn{1}{l|}{0.54} & \multicolumn{1}{l|}{0.93} & \multicolumn{1}{l|}{0.90} & \multicolumn{1}{l|}{0.64} & \multicolumn{1}{l|}{0.64} & \multicolumn{1}{l|}{0.94} & \multicolumn{1}{l|}{0.95} & \multicolumn{1}{l|}{0.93} & 0.99 \\ \hline
\multicolumn{1}{|l|}{\multirow{3}{*}{Univ}} & 20 & \multicolumn{1}{l|}{0.43} & \multicolumn{1}{l|}{0.44} & \multicolumn{1}{l|}{0.89} & \multicolumn{1}{l|}{0.82} & \multicolumn{1}{l|}{0.68} & \multicolumn{1}{l|}{0.68} & \multicolumn{1}{l|}{0.94} & \multicolumn{1}{l|}{0.94} & \multicolumn{1}{l|}{0.53} & 0.53 \\ \cline{2-12} 
\multicolumn{1}{|l|}{} & 100 & \multicolumn{1}{l|}{0.55} & \multicolumn{1}{l|}{0.56} & \multicolumn{1}{l|}{0.97} & \multicolumn{1}{l|}{0.98} & \multicolumn{1}{l|}{0.67} & \multicolumn{1}{l|}{0.67} & \multicolumn{1}{l|}{0.98} & \multicolumn{1}{l|}{0.98} & \multicolumn{1}{l|}{0.76} & 0.79 \\ \cline{2-12} 
\multicolumn{1}{|l|}{} & 500 & \multicolumn{1}{l|}{0.53} & \multicolumn{1}{l|}{0.54} & \multicolumn{1}{l|}{1.00} & \multicolumn{1}{l|}{1.00} & \multicolumn{1}{l|}{0.70} & \multicolumn{1}{l|}{0.70} & \multicolumn{1}{l|}{1.00} & \multicolumn{1}{l|}{1.00} & \multicolumn{1}{l|}{1.00} & 1.00 \\ \hline
\multicolumn{1}{|l|}{\multirow{3}{*}{GNN}} & 20 & \multicolumn{1}{l|}{0.49} & \multicolumn{1}{l|}{0.47} & \multicolumn{1}{l|}{0.88} & \multicolumn{1}{l|}{0.80} & \multicolumn{1}{l|}{0.68} & \multicolumn{1}{l|}{0.68} & \multicolumn{1}{l|}{0.95} & \multicolumn{1}{l|}{0.95} & \multicolumn{1}{l|}{0.53} & 0.52 \\ \cline{2-12} 
\multicolumn{1}{|l|}{} & 100 & \multicolumn{1}{l|}{0.52} & \multicolumn{1}{l|}{0.51} & \multicolumn{1}{l|}{0.97} & \multicolumn{1}{l|}{0.97} & \multicolumn{1}{l|}{0.64} & \multicolumn{1}{l|}{0.64} & \multicolumn{1}{l|}{0.98} & \multicolumn{1}{l|}{0.98} & \multicolumn{1}{l|}{0.69} & 0.60 \\ \cline{2-12} 
\multicolumn{1}{|l|}{} & 500 & \multicolumn{1}{l|}{0.53} & \multicolumn{1}{l|}{0.53} & \multicolumn{1}{l|}{0.99} & \multicolumn{1}{l|}{0.99} & \multicolumn{1}{l|}{0.68} & \multicolumn{1}{l|}{0.68} & \multicolumn{1}{l|}{0.99} & \multicolumn{1}{l|}{0.99} & \multicolumn{1}{l|}{1.00} & 1.00 \\ \hline
\end{tabular}
\caption{Bot detection classification accuracy for the different detectors and synthesis methods using a Closed Set. K=0 implies no use of key-codes when training the classifier and K=1 implies the use of key-codes. The detectors are: One Class Support Vector Machine (OC SVM), Support Vector Machine (SVM), Gaussian Naive Bayes (GNB), Random Forest (RF) and Long short-term memory (LSTM). Accuracy results for evaluation users.}
\label{Table:ResultsClass}
\end{table*} 

\subsection{Results}

The first experiments aim to analyze the capacity of the generation methods to synthesize human-like data (\textbf{O1}) and also to solve the question whether including synthetic data in training can improve the classification results and is therefore useful. For this, we focus on the performances obtained by the one-class classifier (OC SVM) and binary SVM classifier (SVM). The OC SVM is trained using only human samples while the binary SVM is trained using both human and synthetic samples. Both classifiers are evaluated using the same bot and human samples. The results are presented in Table \ref{Table:ResultsClass} (first 4 columns). The OC SVM classifier shows a much lower bot detection accuracy (around 50\%) than the binary SVM (between 77\% and 100\%). On the one hand, the low performance of the OC SVM suggests that synthetic samples present realistic patterns which can not be differentiate from those obtained in real data (using a one-class classification algorithm). On the other hand, the high bot detection accuracy obtained for the binary SVM classifier answers the question about the usefulness of including synthetic samples in training. 

The rest of the columns of the Table \ref{Table:ResultsClass} show the classification accuracy values of the RBF SVM, LSTM, RF, and GNB classifiers trained using the Key-Codes ($f^{5}_{j}$) together with the rest of features ($f^{i}_{j}$, $i\in[1,4]$) (K=1) and without the key-codes (K=0). 

Analyzing again the realism of the synthetic samples (\textbf{O1)} comparing the different synthesis methods, the classification  accuracy of the models trained with the synthetic samples generated with the user-dependent model are lower than those generated with the GNN model (Table \ref{Table:ResultsClass}). Furthermore, the GNN model presents in general lower results than the universal one. A conclusion can be drawn from here: from the way in which the synthetic samples are generated and the models trained, it is more relevant a coherence between the different keystroke time features $f^{i}, i \in [0,4]$ than the coherence between each keystroke time feature  $f^{i}, i \in [0,4]$ and the key-code $f^{i}, i = 5$. .

\begin{table*}[t!]
\centering
\begin{tabular}{ccl|llllllllll|}
\cline{4-13}
\multicolumn{1}{l}{} & \multicolumn{1}{l}{} &  & \multicolumn{10}{c|}{Clasification Model} \\ \cline{1-2} \cline{4-13} 
\multicolumn{2}{|l|}{Gen Model} &  & \multicolumn{2}{c|}{OC SVM} & \multicolumn{2}{c|}{SVM} & \multicolumn{2}{c|}{GNB} & \multicolumn{2}{c|}{RF} & \multicolumn{2}{c|}{LSTM} \\ \hline
\multicolumn{1}{|l|}{Train} & \multicolumn{1}{l|}{Test} & \# Train Subjects & \multicolumn{1}{l|}{K=0} & \multicolumn{1}{l|}{K=1} & \multicolumn{1}{l|}{K=0} & \multicolumn{1}{l|}{K=1} & \multicolumn{1}{l|}{K=0} & \multicolumn{1}{l|}{K=1} & \multicolumn{1}{l|}{K=0} & \multicolumn{1}{l|}{K=1} & \multicolumn{1}{l|}{K=0} & K=1 \\ \hline
\multicolumn{1}{|c|}{\multirow{3}{*}{Univ}} & \multicolumn{1}{c|}{\multirow{3}{*}{GNN}} & 20 & \multicolumn{1}{l|}{0.49} & \multicolumn{1}{l|}{0.48} & \multicolumn{1}{l|}{0.70} & \multicolumn{1}{l|}{0.72} & \multicolumn{1}{l|}{0.63} & \multicolumn{1}{l|}{0.63} & \multicolumn{1}{l|}{0.84} & \multicolumn{1}{l|}{0.82} & \multicolumn{1}{l|}{0.48} & 0.47 \\ \cline{3-13} 
\multicolumn{1}{|c|}{} & \multicolumn{1}{c|}{} & 100 & \multicolumn{1}{l|}{0.54} & \multicolumn{1}{l|}{0.55} & \multicolumn{1}{l|}{0.67} & \multicolumn{1}{l|}{0.70} & \multicolumn{1}{l|}{0.63} & \multicolumn{1}{l|}{0.63} & \multicolumn{1}{l|}{0.68} & \multicolumn{1}{l|}{0.74} & \multicolumn{1}{l|}{0.66} & 0.65 \\ \cline{3-13} 
\multicolumn{1}{|c|}{} & \multicolumn{1}{c|}{} & 500 & \multicolumn{1}{l|}{0.53} & \multicolumn{1}{l|}{0.54} & \multicolumn{1}{l|}{0.59} & \multicolumn{1}{l|}{0.62} & \multicolumn{1}{l|}{0.64} & \multicolumn{1}{l|}{0.64} & \multicolumn{1}{l|}{0.51} & \multicolumn{1}{l|}{0.54} & \multicolumn{1}{l|}{0.99} & 0.99 \\ \hline
\multicolumn{1}{|c|}{\multirow{3}{*}{GNN}} & \multicolumn{1}{c|}{\multirow{3}{*}{Univ}} & 20 & \multicolumn{1}{l|}{0.49} & \multicolumn{1}{l|}{0.48} & \multicolumn{1}{l|}{0.78} & \multicolumn{1}{l|}{0.68} & \multicolumn{1}{l|}{0.65} & \multicolumn{1}{l|}{0.65} & \multicolumn{1}{l|}{0.88} & \multicolumn{1}{l|}{0.89} & \multicolumn{1}{l|}{0.56} & 0.58 \\ \cline{3-13} 
\multicolumn{1}{|c|}{} & \multicolumn{1}{c|}{} & 100 & \multicolumn{1}{l|}{0.53} & \multicolumn{1}{l|}{0.53} & \multicolumn{1}{l|}{0.94} & \multicolumn{1}{l|}{0.91} & \multicolumn{1}{l|}{0.64} & \multicolumn{1}{l|}{0.64} & \multicolumn{1}{l|}{0.91} & \multicolumn{1}{l|}{0.93} & \multicolumn{1}{l|}{0.69} & 0.68 \\ \cline{3-13} 
\multicolumn{1}{|c|}{} & \multicolumn{1}{c|}{} & 500 & \multicolumn{1}{l|}{0.53} & \multicolumn{1}{l|}{0.54} & \multicolumn{1}{l|}{0.97} & \multicolumn{1}{l|}{0.98} & \multicolumn{1}{l|}{0.64} & \multicolumn{1}{l|}{0.64} & \multicolumn{1}{l|}{0.94} & \multicolumn{1}{l|}{0.95} & \multicolumn{1}{l|}{1.00} & 1.00 \\ \hline
\end{tabular}
\caption{Bot detection classification accuracy for the different detectors and synthesis methods using an Open Set. K=0 implies no use of key-codes when training the classifier and K=1 implies the use of key-codes. The detectors are: One Class Support Vector Machine (OC SVM), Support Vector Machine (SVM), Gaussian Naive Bayes (GNB), Random Forest (RF) and Long short-term memory (LSTM). Accuracy results for evaluation users.}
\label{Table:ResultsClassOS}
\end{table*}

The following analysis of the bot detection performance (\textbf{O2}), is divided according to the number of samples available to train the classifiers (Table \ref{Table:ResultsClass}).

\par \textbf{Large (500 subjects)}: 
\textit{Universal and GNN methods}: the results suggest that with enough subjects, perfect classification is achieved using SVM, LSTM and RF. GNB does not achieve high accuracy because this algorithm does not take into account correlations between the different keystroke time features. This also explains why the result is the same with or without the use of key-codes. \textit{User-dependent method}: perfect classification is only achieved with the LSTM classifier and using the key-codes. The LSTM classifier is the one with the highest accuracy when there is enough information to train it. Also, using the key-code information allows to identify better, uncorrelated time features with the key-code. The SVM and RF methods achieve similar performances while the GNB has the lowest accuracy. This is again due to the fact that this method does not take into account correlations between the different keystroke time features. For the same reason it also has a performance similar to the universal and GNN methods.

\par \textbf{Medium (100 subjects)}:
\textit{Universal and GNN methods}: The performance of the LSTM classifier plummets (an average $30$ \%), the top results are achieved with SVM and RF classifiers. This is because these classifiers do not need as much training as LSTM. \textit{User-dependent method}: The classifier with the best accuracy in this case is RF over SVM ($1$\% to $5$\% more). The synthetic and real samples are closer together in the multidimensional space used by SVM as they are more similar to each other, so it needs more training to correctly tune the hyperplane that separates them. In this case also the LSTM classifier performs worse than GNB also because the synthesis is more complex and the classifier needs more training.

\par \textbf{Limited (20 subjects)}: For both universal and user-dependent methods, RF offers the highest performance when there is a high sample sparsity. The RF algorithm based on tree decision favors detection with sparse samples.

The use or not of key-codes does not affect the RF and GNB classifiers at all. The SVM classifier perform worse when the key-codes are included in the feature vector. Note that the text used to generate the bot samples was directly extracted from human samples, therefore, the inclusion of the key-codes did not result in an advantage during the bot detection for this classifier. Nevertheless, the LSTM classifier is capable of associating each key with a time and for this reason it detects impostor samples better in Universal and User-dependent models. In the case of the GNN model, these times have been taken into account to create the synthetic samples and therefore it is more difficult for the classifier to detect them.

The generalization ability of the bot detector was presented in  Table \ref{Table:ResultsClassOS} with a cross database experiment. We focus on the Universal and GNN methods (both methods are user independent so their comparison is fair). In this case, the samples used to train the bot detectors were generated with a method different of the one used to generate the test samples. The results demonstrate that, in general, training with samples synthesized with the GNN model allows to better generalize against unseen synthetic samples of the counter model. 


\begin{table}[t!]
\centering
\begin{tabular}{l|clcl|}
\cline{2-5}
 & \multicolumn{4}{c|}{Gen Model} \\ \cline{2-5} 
 & \multicolumn{1}{l|}{Train} & \multicolumn{1}{l|}{Test} & \multicolumn{1}{l|}{Train} & Test \\ \hline
\multicolumn{1}{|l|}{Method} & \multicolumn{1}{c|}{Univ} & \multicolumn{1}{l|}{GNN} & \multicolumn{1}{c|}{GNN} & Univ \\ \hline
\multicolumn{1}{|l|}{\cite{alamri2022investigation} (Euclidean)} & \multicolumn{2}{c|}{0.49} & \multicolumn{2}{c|}{0.49} \\ \hline
\multicolumn{1}{|l|}{\cite{stefan12} (SVM)} & \multicolumn{2}{c|}{0.62} & \multicolumn{2}{c|}{0.98} \\ \hline
\multicolumn{1}{|l|}{\textbf{Ours} (OCSVM)} & \multicolumn{2}{c|}{0.54} & \multicolumn{2}{c|}{0.54} \\ \hline
\multicolumn{1}{|l|}{\textbf{Ours} (RF)} & \multicolumn{2}{c|}{0.54} & \multicolumn{2}{c|}{0.95} \\ \hline
\multicolumn{1}{|l|}{\textbf{Ours} (GNB)} & \multicolumn{2}{c|}{0.64} & \multicolumn{2}{c|}{0.64} \\ \hline
\multicolumn{1}{|l|}{\textbf{Ours} (LSTM)} & \multicolumn{2}{c|}{\textbf{0.99}} & \multicolumn{2}{c|}{\textbf{1.00}} \\ \hline
\end{tabular}
\caption{Classification accuracy comparison between the proposed approaches and existing methods. The experiments have been carried out assuming a large number of training subjects (500), the use of the key-codes (K=1) and Open Set environment.}
\label{Table:SOTACOMP}
\vspace{-4mm}
\end{table} 

In the last experiment (Table \ref{Table:SOTACOMP}) we compare our classifiers with previous state-of-the-art keystroke bot detection approaches \cite{stefan12, alamri2022investigation}. The features used in \cite{stefan12, alamri2022investigation} were similar to the time features employed in our methods. For a fair comparison, we train and evaluate the methods proposed in \cite{stefan12, alamri2022investigation} with the same synthetic and real samples used in our experiments. This comparison has been carried out assuming a large number of training subjects (500), using key-codes and Open-set environment. The results in the table show that the GNN synthetic samples represents a more difficult challenge for the detectors. The detection performance of these samples varies from $49\%$ to $99\%$.  The results suggest that GNN samples can be used to detect synthetic samples generated with a different synthesizer approach. Our LSTM classifier presents the highest detection performance, achieving a $100\%$ bot detection accuracy for both types of synthetic samples.

\section{Conclusions and Limitations}

In this work we have analyzed the feasibility of using a behavioral trait (dynamic typing) such as passive CAPTCHA where the subject has no need to perform any activity in order for the system to determine if this subject is a bot or a human. 

To train and test the classification models, synthetic samples have been created. We have analyzed three different synthesis methods (Universal, User-dependent, and GNN).

With the synthetic subjects generated with these methods the classification system were trained. We employed different classification algorithms including SVM, RF, GNB, and LSTM network, each one has different behaviour and performance. Depending on the classification system, the generation part has a different performance but with enough training data the classification system is able to perfectly classify between humans and bots. Concluding that the Keystroke-Dynamics can be used as passive CAPTCHA.

Another important result of this work is the proposal of a novel Generative Neural Network. This network allows learning the distribution followed by the different classes within a data set. It is a pioneering network both for its architecture and for the way it learns from the data, with a loss function that evaluates the distribution. In addition, in terms of concept, it is also innovative, normally neural networks have a set of inputs that belong to different classes, but each class is a value. In this case each input value belongs to a class and this class is a distribution.

The utilities of this network are many, whenever you want to learn a distribution or focus the learning of a network on distributions instead of individual values. The potential of this network lies in using the network as a unit and creating a network formed by these units, in this way one could learn complex functions (even non-linear) and have a non-deterministic network in classification. 

The main line of future work is a system that presents both intra-user dependencies and key dependencies. To this end, different generative systems can be trained for different subjects (Generative user-dependent) or a certain correlation between the different keystroke time features can be included in the learning of the generative network itself.

\section*{Acknowledgment}

This work has been supported by projects: TRESPASS-ETN (MSCA-ITN-2019-860813) and BBforTAI (PID2021-127641OB-I00 MICINN/FEDER). The work of D. deAlcala is supported by a FPU Fellowship (FPU21/05785) by the Spanish MIU.

{\small
\bibliographystyle{ieee_fullname}
\bibliography{egbib}

\begin{thebibliography}{10}\itemsep=-1pt

\bibitem{acien22mouse}
A. Acien, A. Morales, J. Fierrez, and R. Vera-Rodriguez.
\newblock {BeCAPTCHA-Mouse: Synthetic Mouse Trajectories and Improved Bot
  Detection}.
\newblock {\em Pattern Recognition}, 127:108643, 2022.

\bibitem{acien21mobile}
A. Acien, A. Morales, J. Fierrez, R. Vera-Rodriguez, and O. Delgado-Mohatar.
\newblock {BeCAPTCHA: Behavioral Bot Detection using Touchscreen and Mobile
  Sensors benchmarked on HuMIdb}.
\newblock {\em Engineering Applications of Artificial Intelligence}, 98:104058,
  2021.

\bibitem{acien22transactions}
A. Acien, A. Morales, John~V. Monaco, R. Vera-Rodríguez, and Julian Fierrez.
\newblock {TypeNet}: Deep learning keystroke biometrics.
\newblock {\em IEEE Transactions on Biometrics, Behavior, and Identity
  Science}, 4(1):57--70, 2022.

\bibitem{al2008detecting}
Yousof Al-Hammadi and Uwe Aickelin.
\newblock Detecting bots based on keylogging activities.
\newblock In {\em 2008 Third International Conference on Availability,
  Reliability and Security}, pages 896--902, 2008.

\bibitem{alamri2022investigation}
Emtethal~K Alamri, Abdullah~M Alnajim, and Suliman~A Alsuhibany.
\newblock Investigation of using captcha keystroke dynamics to enhance the
  prevention of phishing attacks.
\newblock {\em Future Internet}, 14(3):82, 2022.

\bibitem{ali2016keystroke}
Md~Liakat Ali, Kutub Thakur, Charles~C Tappert, and Meikang Qiu.
\newblock Keystroke biometric user verification using hidden markov model.
\newblock In {\em 2016 IEEE 3rd International Conference on Cyber Security and
  Cloud Computing (CSCloud)}, pages 204--209, 2016.

\bibitem{ayotte2020fast}
Blaine Ayotte, Mahesh Banavar, Daqing Hou, and Stephanie Schuckers.
\newblock Fast free-text authentication via instance-based keystroke dynamics.
\newblock {\em IEEE Transactions on Biometrics, Behavior, and Identity
  Science}, 2(4):377--387, 2020.

\bibitem{bergadano2002user}
Francesco Bergadano, Daniele Gunetti, and Claudia Picardi.
\newblock User authentication through keystroke dynamics.
\newblock {\em ACM Transactions on Information and System Security (TISSEC)},
  5(4):367--397, 2002.

\bibitem{chatgpt}
Vallance Chris.
\newblock Chatgpt: New {AI} chatbot has everyone talking to it.
\newblock {\em https://www.bbc.com/news/technology-63861322}, BBC, 7 December
  2022.

\bibitem{2022_ACM_Privacy_Paula_Giuseppe}
Paula Delgado-Santos, Giuseppe Stragapede, Ruben Tolosana, Richard Guest,
  Farzin Deravi, and Ruben Vera-Rodriguez.
\newblock A survey of privacy vulnerabilities of mobile device sensors.
\newblock {\em ACM Computing Surveys}, 2022.

\bibitem{dhakal2018observations}
Vivek Dhakal, Anna~Maria Feit, Per~Ola Kristensson, and Antti Oulasvirta.
\newblock Observations on typing from 136 million keystrokes.
\newblock In {\em Proceedings of the 2018 CHI Conference on Human Factors in
  Computing Systems}, pages 1--12, 2018.

\bibitem{gaines1980}
R. Gaines, S. Press, W. Lisowski, and N. Shapiro.
\newblock Authentication by keystroke timing : some preliminary results.
  rapport technique.
\newblock {\em Rand Corporation}, 1980.

\bibitem{gamboa2007webbiometrics}
H Gamboa, ALN Fred, and AK Jain.
\newblock Webbiometrics: User verification via web interaction.
\newblock In {\em 2007 Biometrics Symposium}, pages 1--6, 2007.

\bibitem{webcyber}
Radauskas Gintaras.
\newblock {AI}-enabled cyberattacks might become norm in next five years.
\newblock {\em https://cybernews.com/news/ai-enabled-cyberattacks-new-norm/},
  Cybernews, 15 December 2022.

\bibitem{gonzalez2022towards}
Nahuel Gonz{\'a}lez, Enrique~P Calot, Jorge~S Ierache, and Waldo Hasperu{\'e}.
\newblock Towards liveness detection in keystroke dynamics: Revealing synthetic
  forgeries.
\newblock {\em Systems and Soft Computing}, 4:200037, 2022.

\bibitem{grover2018flow}
Aditya Grover, Manik Dhar, and Stefano Ermon.
\newblock Flow-gan: Combining maximum likelihood and adversarial learning in
  generative models.
\newblock In {\em Proceedings of the AAAI conference on artificial
  intelligence}, volume~32, 2018.

\bibitem{jain16}
Anil Jain, Karthik Nandakumar, and Arun Ross.
\newblock 50 years of biometric research: Accomplishments, challenges, and
  opportunities.
\newblock {\em Pattern Recognition Letters}, 79:80--105, 2016.

\bibitem{kim2012robust}
JooSeuk Kim and Clayton~D Scott.
\newblock Robust kernel density estimation.
\newblock {\em The Journal of Machine Learning Research}, 13(1):2529--2565,
  2012.

\bibitem{liu2021density}
Qiao Liu, Jiaze Xu, Rui Jiang, and Wing~Hung Wong.
\newblock Density estimation using deep generative neural networks.
\newblock {\em Proceedings of the National Academy of Sciences},
  118(15):e2101344118, 2021.

\bibitem{mhenni2019vulnerability}
Abir Mhenni, Denis Migdal, Estelle Cherrier, Christophe Rosenberger, and Najoua
  Essoukri~Ben Amara.
\newblock Vulnerability of adaptive strategies of keystroke dynamics based
  authentication against different attack types.
\newblock In {\em 2019 International Conference on Cyberworlds (CW)}, pages
  274--278, 2019.

\bibitem{migdal2018analysis}
Denis Migdal and Christophe Rosenberger.
\newblock Analysis of keystroke dynamics for the generation of synthetic
  datasets.
\newblock In {\em 2018 International Conference on Cyberworlds (CW)}, pages
  339--344, 2018.

\bibitem{migdal2019statistical}
Denis Migdal and Christophe Rosenberger.
\newblock Statistical modeling of keystroke dynamics samples for the generation
  of synthetic datasets.
\newblock {\em Future Generation Computer Systems}, 100:907--920, 2019.

\bibitem{monaco2016robust}
John~V Monaco.
\newblock Robust keystroke biometric anomaly detection.
\newblock {\em arXiv preprint arXiv:1606.09075}, 2016.

\bibitem{monaco2015spoofing}
John~V Monaco, Md~Liakat Ali, and Charles~C Tappert.
\newblock Spoofing key-press latencies with a generative keystroke dynamics
  model.
\newblock In {\em 2015 IEEE 7th international conference on biometrics theory,
  applications and systems (BTAS)}, pages 1--8, 2015.

\bibitem{morales2022setmargin}
Aythami Morales, Julian Fierrez, Alejandro Acien, Ruben Tolosana, and Ignacio
  Serna.
\newblock Setmargin loss applied to deep keystroke biometrics with circle
  packing interpretation.
\newblock {\em Pattern Recognition}, 122:108283, 2022.

\bibitem{morales16keystroke}
A. Morales, J. Fierrez, R. Tolosana, J. Ortega-Garcia, J. Galbally, M.
  Gomez-Barrero, A. Anjos, and S. Marcel.
\newblock Keystroke biometrics ongoing competition.
\newblock {\em IEEE Access}, page 7736–7746, 2016.

\bibitem{murphy2017shared}
Christopher Murphy, Jiaju Huang, Daqing Hou, and Stephanie Schuckers.
\newblock Shared dataset on natural human-computer interaction to support
  continuous authentication research.
\newblock In {\em 2017 IEEE International Joint Conference on Biometrics
  (IJCB)}, pages 525--530, 2017.

\bibitem{tolo_handbook_deepfakes}
Christian Rathgeb, Ruben Tolosana, Ruben Vera-Rodriguez, and Christoph Busch.
\newblock {\em Handbook of Digital Face Manipulation and Detection: From
  DeepFakes to Morphing Attacks}.
\newblock Springer, 2022.

\bibitem{singh2019issues}
Manmeet Singh, Maninder Singh, and Sanmeet Kaur.
\newblock Issues and challenges in dns based botnet detection: A survey.
\newblock {\em Computers \& Security}, 86:28--52, 2019.

\bibitem{stefan12}
D. Stefan, S. Xun, and D. Yao.
\newblock {Robustness of keystroke-dynamics based biometrics against synthetic
  forgeries}.
\newblock {\em Computers \& Security}, pages 109--121, 2012.

\bibitem{2023_FG_KeystrokeTransformers_Giuseppe}
Giuseppe Stragapede, Paula Delgado-Santos, Ruben Tolosana, Ruben
  Vera-Rodriguez, Richard Guest, and Aythami Morales.
\newblock Mobile keystroke biometrics using transformers.
\newblock In {\em Proc. IEEE Intl. Conf. on Automatic Face and Gesture
  Recognition (FG)}, 2023.

\bibitem{uria2016neural}
Benigno Uria, Marc-Alexandre C{\^o}t{\'e}, Karol Gregor, Iain Murray, and Hugo
  Larochelle.
\newblock Neural autoregressive distribution estimation.
\newblock {\em The Journal of Machine Learning Research}, 17(1):7184--7220,
  2016.

\bibitem{wang2023neural}
Chengyi Wang, Sanyuan Chen, Yu Wu, Ziqiang Zhang, Long Zhou, Shujie Liu, Zhuo
  Chen, Yanqing Liu, Huaming Wang, Jinyu Li, et~al.
\newblock Neural codec language models are zero-shot text to speech
  synthesizers.
\newblock {\em arXiv preprint arXiv:2301.02111}, 2023.

\bibitem{zhu2017unpaired}
Jun-Yan Zhu, Taesung Park, Phillip Isola, and Alexei~A Efros.
\newblock Unpaired image-to-image translation using cycle-consistent
  adversarial networks.
\newblock In {\em Proceedings of the IEEE international conference on computer
  vision}, pages 2223--2232, 2017.

\end{thebibliography}
}

\end{document}